# Performance Analysis of NDT-based Graph SLAM for Autonomous Vehicle in Diverse Typical Driving Scenarios of Hong Kong

**Weisong Wen, Li-Ta Hsu \*, Guohao Zhang**

Department of Mechanical Engineering, The Hong Kong Polytechnic University, Hong Kong SAR, China; 17902061r@connect.polyu.hk (W.W.)

\* Correspondence: lt.hsu@polyu.edu.hk

**Abstract:** Robust and lane-level positioning is essential for autonomous vehicles. As an irreplaceable sensor, LiDAR can provide continuous and high-frequency pose estimation by means of mapping, on condition that enough environment features are available. The error of mapping can accumulate over time. Therefore, LiDAR is usually integrated with other sensors. In diverse urban scenarios, the environment feature availability relies heavily on the traffic (moving and static objects) and the degree of urbanization. Common LiDAR-based SLAM demonstrations tend to be studied in light traffic and less urbanized area. However, its performance can be severely challenged in deep urbanized cities, such as Hong Kong, Tokyo, and New York with dense traffic and tall buildings. This paper proposes to analyze the performance of standalone NDT-based graph SLAM and its reliability estimation in diverse urban scenarios to further evaluate the relationship between the performance of LiDAR-based SLAM and scenario conditions. The normal distribution transform (NDT) is employed to calculate the transformation between frames of point clouds. Then, the LiDAR odometry is performed based on the calculated continuous transformation. The state-of-the-art graph-based optimization is used to integrate the LiDAR odometry measurements to implement optimization. The 3D building models are generated and the definition of the degree of urbanization based on Skyplot is proposed. Experiments are implemented in different scenarios with different degrees of urbanization and traffic conditions. The results show that the performance of the LiDAR-based SLAM using NDT is strongly related to the traffic condition and degree of urbanization. The best performance is achieved in the sparse area with normal traffic and the worse performance is obtained in dense urban area with 3D positioning error (summation of horizontal and vertical) gradients of 0.024 m/s and 0.189 m/s, respectively. The analyzed results can be a comprehensive benchmark for evaluating the performance of standalone NDT-based graph SLAM in diverse scenarios which is significant for multi-sensor fusion of autonomous vehicle.

**Keywords:** Localization; NDT; Graph SLAM; LiDAR; Autonomous Vehicle

## 1. Introduction

Autonomous vehicle [1,2] is well believed to be the next revolutionary technology changing people's lives in many ways. For fully autonomous vehicles, localization is one of the key parts because accurate and robust positioning is the basics of further perception and path planning missions for autonomous driving. The most promising solution to provide the globally referenced positioning is sensor integration of global navigation satellites system (GNSS), light detection and ranging (LiDAR), inertial measurement unit (IMU) and high definition map (HD map). Currently, this solution can provide satisfactory performance in suburban areas [3-9]. However, the performance of the integration solution can be severely challenged in the deep urbanized area, such as Hong Kong, Tokyo, and New York. Firstly, the accuracy of GNSS positioning can be decreased to 50 meters [10,11], due to the blockage, reflection, and diffraction from buildings and moving objects.

Moreover, the uncertainty of GNSS positioning, which is significant for sensor fusion, is difficult to model. The IMU can drift over time due to the dense traffic congestions. The matching between the real-time point clouds from LiDAR and offline point clouds from HD Map can also be challenged due to the excessive moving objects, changeable city structures, and environment feature availability. Simultaneous localization and mapping (SLAM) [12,13] is a significant method to provide positioning service based on mapping of point clouds. The accuracy of SLAM-based positioning relies heavily on the mapping between point clouds. In other words, the performance of SLAM is strongly related to the environment. Thus, this paper proposes to evaluate the performance of NDT-based graph SLAM in diverse urban scenarios to further study the relationship, between the performance of SLAM and environment conditions.

Numerous studies [13-18] are conducted in the past decades on the LiDAR-based SLAM. The main principle of LiDAR-based SLAM is to continuously track the transformation between successive frames of point clouds. In this case, the performance of SLAM relies heavily on the accuracy of the mapping-based transformation. The LOAM [18] can obtain low drift positioning when satisfactory environment features are available, such as the planes and edges. However, the performance of this algorithm can be severely degraded in dense urban, due to the excessive moving objects. On one hand, the positioning accuracy cannot be guaranteed with too much dynamic plane features from traffic. On the other hand, the LOAM algorithm did not propose an effective method to model the uncertainty of LiDAR-based positioning. The submap concept is proposed in [13] and the real-time loop closure detection is achieved. However, the uncertainty is not well modeled, and the performance of this algorithm relies heavily on the additional sensors, such as the IMU. The mapping solution, normal distribution transform (NDT), based on the normal distribution transform is proposed in [19]. This method innovatively employs the normal distribution transform to transfer the mapping process into probabilistic continuous functions. However, this method also cannot effectively model the positioning uncertainty caused by moving objects. According to the previous research, the uncertainty caused by the moving objects are not well modeled [20]. To reduce the drift of LiDAR-based positioning, the loop closure detection [14] algorithm is proposed to mitigate the global positioning error. The main idea for loop detection is to identify the two similar pose that the vehicle has gone through. Then the overall correction of the poses is obtained based on the loop closure to improve the accuracy. However, the loop closure is subjected to the availability of closed loop.

Both the LOAM and the NDT can be used to calculate the transformation between the consecutive frames of point clouds. The LiDAR odometry can be obtained by tracking the continuous transformations. To integrate the LiDAR odometry and other sensors, a sensor fusion framework is needed. Based on the principle of the sensor integration, the sensor integration methods can be divided into two groups, the filtering-based and the smoothing-based integration. The symbolic filtering-based sensor integration method is the Bayes filter, including Kalman filter [21,22], information filter [16,23,24] and particle filter [17,25,26]. The Bayes filter-based sensor integration estimates the current state only based on current observation and the previous state estimation, abandoning all the states before the previous states [27]. This is because of the assumption of the first order of the Markov model which is one of the key assumptions of the Bayes filter. Conversely, the smoothing approaches [28-31] estimate the pose and map by considering the full sets of measurements between the first epoch and the current epoch. The most well-known smoothing method is the graph-based SLAM [12]. However, no matter which mapping method is chosen, the accuracy of LiDAR-based positioning is significant for final sensor integration.

Our team aims to develop autonomous driving vehicles to facilitate the next generation of the intelligent transportation system of Hong Kong. Accurate and robust localization service is the basis. This paper extensively analyzes the performance of NDT-based graph SLAM in diverse urban scenarios. This paper firstly employs the NDT to calculate the transformation between two consecutive frames. Then, the graph optimization is used to optimize all the LiDAR odometry measurements from the first epoch to current epoch. Moreover, this paper estimates the uncertainty of the LiDAR odometry in terms of the degree of matching, number of iterations and time used for NDT optimization. This covariance estimation solution is available in point cloud library (PCL).

The main contributions of this paper are listed as follows:
(1) This paper proposes to generate the 3D building models of the tested area to define the degree of urbanization of the given scenario. The Skyplot is generated as an indicator of the degree of urbanization and the corresponding definition is presented. The classification criteria of different urban scenarios are proposed using Skyplot features.
(2) The multi-sensor integrated localization solution [7,8] tend to evaluate the performance in sparse scenarios with friendly traffic. This paper evaluates the performance of NDT-based graph SLAM in diverse urban scenarios, with different traffic conditions and degree of urbanization.
(3) This paper qualitatively analyzes the relationship between the performance of NDT-based graph SLAM and the traffic conditions and degree of urbanization. The evaluated results related to the traffic and performance of LiDAR-based positioning can be a useful basic work for further mitigating the effects of traffic and urbanization to improve the performance of LiDAR-based positioning.

The rest of the paper is structured as follows. The transformation calculation based on LiDAR is presented in Section 2. The graph-based SLAM is introduced in Section 3 before the experiment evaluation is given in Section 4. Finally, the conclusion and future work are presented in Section 5.

## 2. The transformation from LiDAR-based Mapping

*2.1 Transformation Calculation*

The principle of LiDAR odometry [18] is to track the transformation between two successive frames of point clouds by matching the two frames of point clouds called reference point cloud and input point cloud in this paper. The matching process is also called point cloud registration. The objective of point cloud registration is to obtain the optimal transformation matrix to match or align the reference and the input point clouds. The most well-known and conventional method of point cloud registration is the iterative closest point (ICP) [32] method. The ICP is a straightforward method to calculate the transformation between two consecutive scans by iteratively searching pairs of nearby points in the two scans and minimizing the sum of all point-to-point distances. The objective function can be expressed as follow [33]:

$$C(\hat{\mathbf{R}}, \hat{\mathbf{T}}) = \arg\min \sum_{i=1}^{N} ||(\mathbf{R}p_i + \mathbf{T}) - q_i||^2 \qquad (1)$$

where the $N$ indicates the number of points in one scan. $\mathbf{R}$ and $\mathbf{T}$ indicates the rotation and translation matrix respectively, to transform the input point cloud ($p_i$) into the reference point cloud ($q_i$). Objective function $C(\hat{\mathbf{R}}, \hat{\mathbf{T}})$ indicates the transformation error.

The NDT, modeling the points based on Gaussian distribution, is the other method to align two consecutive scans. The NDT innovatively divides the point cloud space into cells and each cell is continuously modeled by a Gaussian distribution. In this case, the discrete point clouds are transformed into successive continuous functions. In this paper, the NDT is chosen as the point cloud registration method for LiDAR odometry. Assuming that the transformation between two consecutive frames of point clouds can be expressed as $\mathbf{M} = [T_x \quad T_y \quad T_z \quad R_x \quad R_y \quad R_z]^T$. The process of calculating the relative pose between the reference and the input point clouds is listed as follows:

1) Normal distribution transform: fetch all the points $x_{i=1...n}$ contained in 3D the cell.

   Calculate the mean among all the points, $q = \frac{1}{n}\sum_i x_i$.

   Calculate the covariance matrix $\boldsymbol{\mu}$,

$$\boldsymbol{\mu} = \frac{1}{n}\sum_i (x_i - q)(x_i - q)^T \qquad (2)$$

2) The matching score is modeled as:

$$f(p) = -score(p) = \sum_i \exp\left(-\frac{(x_i'-q_i)^T \mu_i^{-1}(x_i'-q_i)}{2}\right) \quad (3)$$

where $x_i$ indicates the points in the current frame of the scan. $x_i'$ denotes the point in the previous scan mapped from the current frame using the **M**. $q_i$ and $\mu_i$ indicate the mean and the covariance of the corresponding normal distribution to point $x_i'$ in the NDT of the previous scan.

3) Update the pose using the Quasi-Newton method using the objective function to minimize the score $f(p)$.

4) Repeat the steps 2) and 3) until the convergence is achieved.

With all the points in one frame of point clouds being modeled as cells, the objective of the optimization for NDT is to match current cells into the previous cells with the highest probability. The optimization function $f(p)$ can be found in [15]. For each cell containing several points, the corresponding covariance matrix can be calculated and represented by $\Sigma$. The shape (circle, plane or linear) of the cell is indicated by the relations between the three eigenvalues of the covariance matrix.

*2.2 Uncertainty Estimation of Transformation*

As the transformation calculation is not always accurate, the uncertainty is a significant parameter to model the possible error range of the transformation. The uncertainty is modeled as the covariance. The associated covariance is significant for further sensor integration. During the NDT process, the covariance of the transformation calculation is related to the environment condition and the similarity between the consecutive frames of point clouds. In the graph-based optimization which will be presented in Section 3, the covariance is indicated as the inverse of the information matrix $\mathbf{\Omega}_{ij}$. This means the weight of constraint between node *i* and *j*.

In this paper, we propose to use the method in [34] to calculate the covariance of transformation calculation based on the following three items:

(1) The degree of matching between the two consecutive frames of point clouds.
(2) The used time to complete the transformation calculation.
(3) The iteration number used to make the Quasi-Newton method converge.

For each matching process between reference point cloud *i* and input point cloud *j*, we model the degree of matching as:

$$U_{delta,ij} = \frac{1}{n}\sum_{k=1}^{n}\sqrt{(x_{delta,k})^2 + (y_{delta,k})^2 + (z_{delta,k})^2} \quad (4)$$

where the $U_{delta,ij}$ represents the degree of matching between the input and the reference point clouds. $n$ represents the number of the points in input point cloud. $x_{delta,k}$ indicates the positional difference in $x$ axis between an input point and a reference point after the convergence of NDT is obtained. $y_{delta,k}$ and $z_{delta,k}$ indicate the positional differences in $x$ and $y$ axis respectively. If the time used to make the NDT converge is $t_c$, the uncertainty related to the time is denoted as:

$$U_{t,ij} = C_t t_c \quad (5)$$

where the $C_t$ is the coefficient and is heuristically determined. If the iterations number used to make the NDT converge is $N_c$, the uncertainty related to $N_c$ is represented as:

$$U_{N,ij} = C_N N_c \quad (6)$$

where the $C_N$ is the coefficient and is heuristically determined. Thus, for each match between a reference point cloud and an input point cloud, the total uncertainty of the transformation calculation is denoted as $U_{ij}$:

$$U_{ij} = U_{delta,ij} + U_{t,ij} + U_{N,ij} \quad (7)$$

the information matrix $\mathbf{\Omega}_{ij}$ can be expressed as:

$$\mathbf{\Omega}_{ij} = \begin{bmatrix} \Omega_{ij}^p & 0 \\ 0 & \Omega_{ij}^r \end{bmatrix} \quad (8)$$

$$\Omega_{ij}^p = \mathbf{I}/(C_p^2 U_{ij}) \tag{9}$$
$$\Omega_{ij}^r = \mathbf{I}/(C_r^2 U_{ij}) \tag{10}$$

where $\mathbf{I}$ am a unit matrix, $C_p^2$ and $C_r^2$ are heuristically determined coefficients for adjusting the covariance of translation and rotation, respectively. In this case, the covariance of the transformation calculation is correlated with the degree of matching, time used for the NDT convergence and times of iteration.

## 3. Graph-based SLAM

This section presents the graph optimization. Pose graph optimization is to construct all the measurements into a graph as constraints and calculate the best set of poses by solving a non-linear optimization problem [12]. In this paper, the constraints are provided by the continuous LiDAR odometry. To implement the graph-based integration optimization, two steps are needed, the graph generation and graph optimization.

### 3.1 Graph Generation

The graph consists of edges and vertices [12]. As shown in Figure. 1, edges are indicated by observation measurements from the LiDAR odometry. The $x_i$ represents pose estimation that included the position and orientation. $e_{ij}$ indicates the error function evaluating the difference between the estimated state and the observed pose measurements. $z_{ij}$ represents the observation and the $\hat{z}_{ij}$ indicates the expected observation. The blue circles and lines represent the nodes and the edges respectively, which is provided by LiDAR odometry.

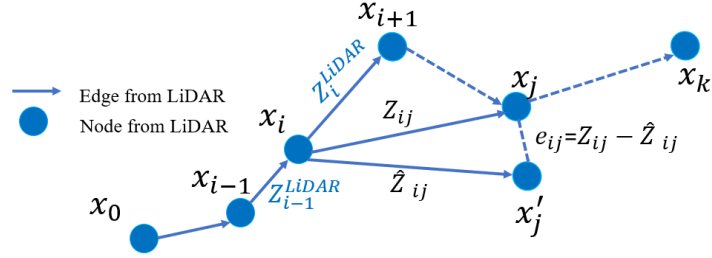

**Figure 1.** Demonstration of the graph generation based on LiDAR odometry.

### 3.2 Graph Optimization

The graph optimization takes all the constraints into a non-linear optimization problem. The main feature of the graph-based optimization is that all the observation measurements are considered. The optimization form is shown as following [35]:

$$F(\mathrm{x}) = \sum_{i,j} e(\mathrm{x}_i, \mathrm{x}_j, \hat{z}_{ij})^T \Omega_{ij} e(\mathrm{x}_i, \mathrm{x}_j, \hat{z}_{ij}) \tag{11}$$

where $F(\mathrm{x})$ is the optimization function which is the sum errors of all the edges. The $\Omega_{ij}$ is the information matrix indicating the importance of each constraint in the global graph optimization. The information matrix is the inverse of the covariance matrix estimated in Section 2. The final solution of this optimization is the $x^*$ which satisfied the following function:

$$x^* = \mathrm{argmin} F(x) \tag{12}$$

Thus, the optimization lies in solving the equation above to obtain the optimal $x^*$. We can see from the optimization form $F(\mathrm{x})$, the covariance of the measurements from LiDAR odometry is represented by the information matrix $\Omega_{ij}$. If the covariance of each measurement is not properly solved, the global optimization will be deflected resulting in the erroneous final pose sets. In other words, the covariance is significant for the performance of the graph optimization.

## 4. Experimental Evaluation

To evaluate the performance of NDT-based graph SLAM in diverse urban scenarios, experiments are conducted in 3 different scenarios with different traffic conditions. The environment features of 3 scenes are shown as following:
(1) Sparse area: (a) Sparse area with normal traffic. (b) The sparse area with dense traffic.
(2) Sub-urban area: (a) Sub-urban area with normal traffic. (b) The sub-urban area with dense traffic. (presented in the appendix)
(3) Dense urban area: (a) Dense urban area with normal traffic. (b) The dense urban area with dense traffic.

**Definition of the degree of urbanization:** The level of urbanization increases from scenes (1) to (3). There is almost no effective and existing way to model the degree of urbanization regarding to the autonomous vehicle. The degree of urbanization is extensively discussed in the GNSS field, as buildings [36,37] and moving objects [20] can have significant effects on the accuracy of GNSS solutions. The Skyplot [38] is employed to represent the satellite visibility (LOS: line-of-sight, NLOS: non-line-of-sight) by project both the 3D building models and satellites into the Skyplot coordinate system. Inspired by this, we propose to use the mean mask elevation angle of a Skyplot as an indicator of urbanization. As the building models information are available in Google Maps for research purpose, we construct the 3D building models of the experiment scene shown in Figure 2.

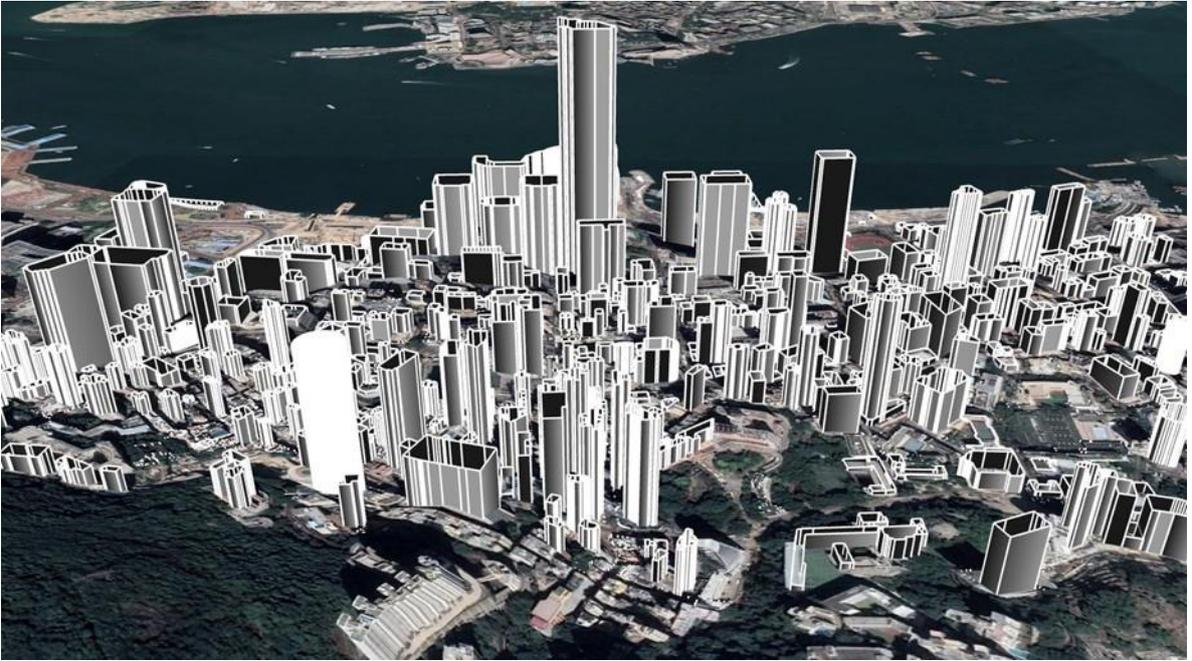

**Figure 2.** 3D building models of tested area.

For a dense urban scenario, the 3D building models generated in a street is shown on the left side in Figure 3. Assuming the vehicle is at the position shown in Figure 3 with 3D LiDAR sensor being installed on the top. By connecting a line from the vehicle and building roof, we can denote the mask elevation angle and azimuth angle as α and θ, respectively. Assume that at an azimuth angle $\theta_i$, the distance between the vehicle and the building is $W$ and building height is $H$, we can calculate the corresponding elevation angle $\alpha_i$ as:

$$\alpha_i = \operatorname{atan}(\frac{H}{W}) \qquad (13)$$

By traversing all the azimuth angles from 0° to 360° with an angular resolution of 1°, we can obtain the Skyplot [38] (shown in right side of Figure 3) regarding the given vehicle position. The inner circle indicates a different elevation angle. The "N" means north of earth. The shaded area means blockage from buildings. Taller buildings cause more blockage, which introduces a higher degree of urbanization.

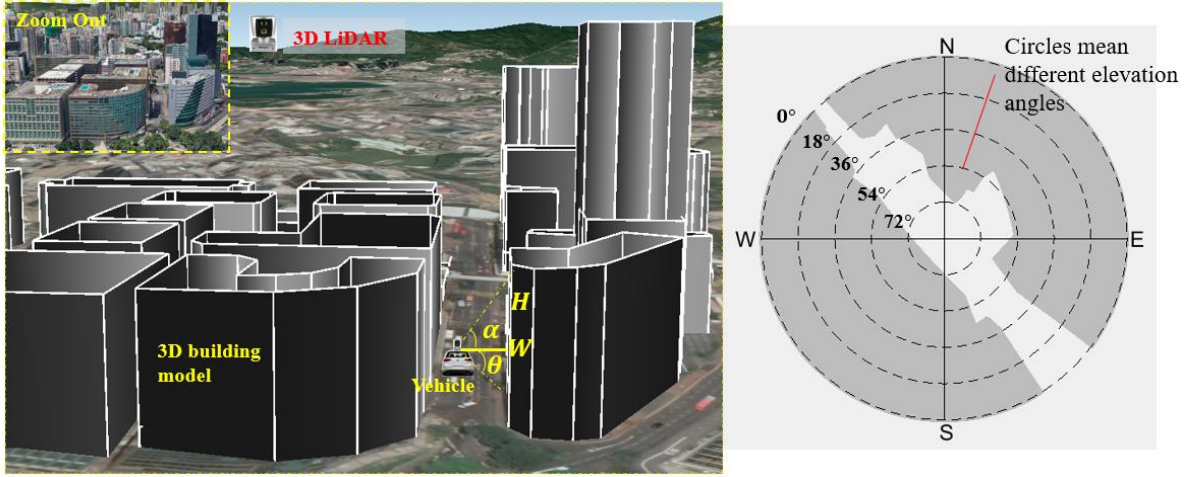

**Figure 3.** Skyplot generation based on 3D building models.

Therefore, we propose to calculate the mean elevation mask angle as a quantitative indicator of the degree of urbanization of the given scenario as follows:

$$\beth_{urban} = \sum_{i=1}^{360} \alpha_i / 360 \qquad (14)$$

In this case, the degree of urbanization of a given scenario is represented by $\beth_{urban}$. The $\beth_{urban}$ for the scenario shown in Figure 3 is 55.32°. We propose to define the degree of urbanization using the rules in TABLE 1. Therefore, the $\beth_{urban}$ for the scenario in Figure 3 satisfies the dense urban area condition.

**Table 1.** Definition of the degree of urbanization based on $\beth_{urban}$

|  | Sparse area | Sub-urban area | Dense urban area |
|---|---|---|---|
| $\beth_{urban}$ | 0°~15° | 15°~46° | > 46° |

**Definition of traffic conditions:** We define that the normal traffic means the common traffic condition density in urban with approximately 2~5 vehicles surrounding the ego-vehicle (shown in the left of Figure 4). The dense traffic indicates that there are numerous moving objects on the roads. For example, in rush hours with approximately 8~12 vehicles surrounding the ego-vehicle (shown in the right of Figure 4).

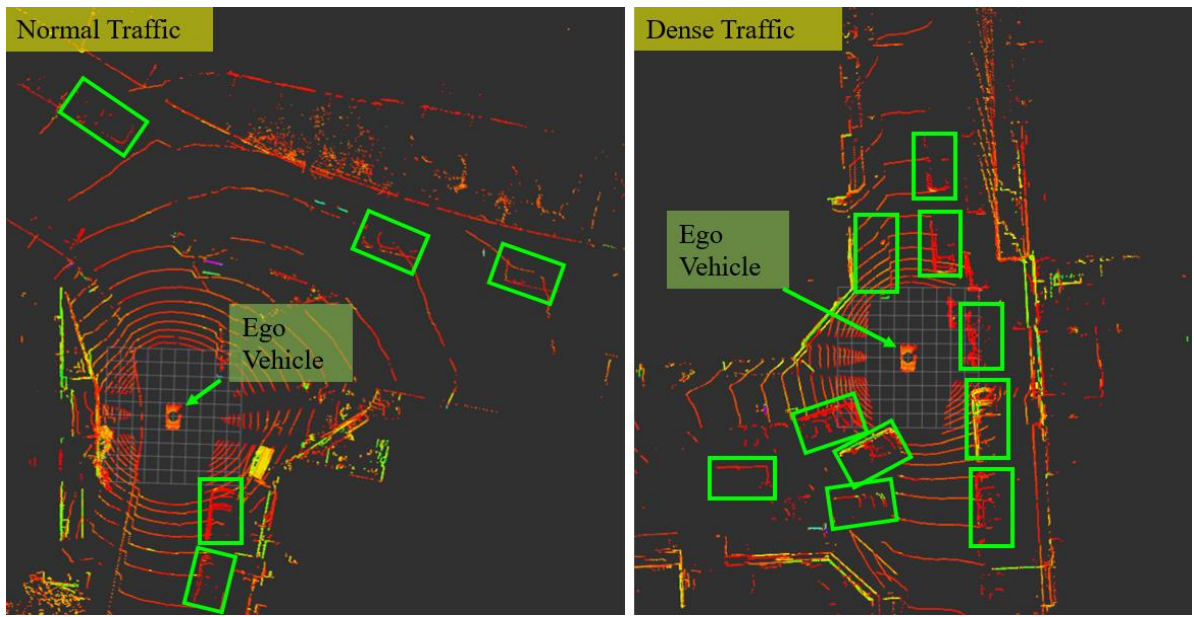

**Figure 4.** Scenarios with different traffic conditions. The green boxes indicate the surrounding dynamic vehicles. The left figure shows the scenario with normal traffic and right figure with dense traffic.

*4.1 Experimental Setup*

3D LiDAR sensor, Velodyne 32, is employed to provide the real-time point clouds scanned from the surroundings. 3D LiDAR is installed on the top of a vehicle during the experiment which can be seen in Figure 5. The LiDAR coordinate system is shown in Figure 5 with *x*-axis pointing back of the vehicle. The integrated navigation system (NovAtel SPAN-CPT, RTK/INS integrated navigation system with fiber optics gyroscopes) based on local ENU [39] coordinate system is used to provide ground truth. The coordinate system of LiDAR and SPAN-CPT is calibrated at the beginning of the tests.

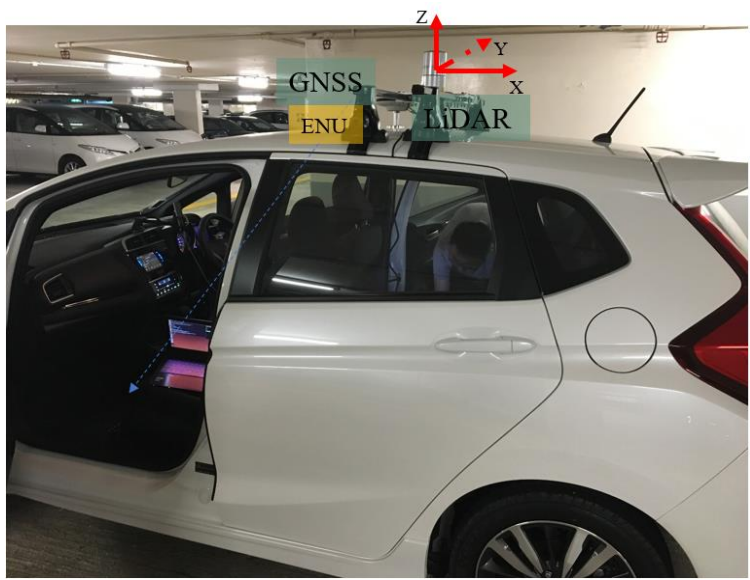

**Figure 5.** Sensors setup of the vehicle: 3D LiDAR sensor is installed on the top of the vehicle. GNSS RTK/INS integrated navigation system is installed on the top of the vehicle next to 3D LiDAR. The GNSS RTK/INS in ENU coordinate system is used to provide the ground truth of vehicle's position.

The evaluated items in the following experiments include positioning error of NDT-based graph SLAM and reliability estimation error. Regarding the total positioning error $\epsilon_{SLAM}$, it is calculated by:

$$\epsilon_{SLAM}= \sqrt{((E^e_{slam} - E^e_{GT})^2 + (N^n_{slam} - N^n_{GT})^2 + (U^u_{slam} - U^u_{GT})^2)} \qquad (15)$$

where $E^e_{slam}$, $N^n_{slam}$ and $U^u_{slam}$ denote the estimated position in east, north and upward axis of ENU coordinate system. The $E^e_{GT}$, $E^n_{GT}$ and $E^u_{GT}$ denote the ground truth (by GNSS RTK/INS integrated system) in east, north and upward axis of ENU coordinate system. Assuming that the heading angle of vehicle relative to north of the earth is denoted as $R_U$ in radian, then the positioning error $\epsilon^{longitudinal}_{SLAM}$ of a vehicle in longitudinal direction can be calculated by projecting $\epsilon_{SLAM}$ into the direction of heading ($R_U$). Similarly, the error in the lateral direction can also be calculated by projecting $\epsilon_{SLAM}$ into the direction normal to heading direction ($R_U$).

The reliability is calculated as $C_p\sqrt{U_{ij}}$. The ground truth for reliability estimation is the actual total positioning error $\epsilon_{SLAM}$. The objective of reliability estimation is to obtain a smallest circle that can cover $\epsilon_{SLAM}$ to represent the uncertainty of a given positioning result.

*4.2 Experiment in Sparse Area*

4.2.1 Experiment 1: Performance Evaluation of NDT-based Graph SLAM in Sparse Area with Normal Traffic

In this experiment, the scenario is shown in the top panel of Figure 6. The overall drive of vehicle lasts about 395 seconds in a sparse area with normal traffic. The height of the surrounding buildings is about 5~10 meters high and the width of the streets is approximately 16 meters. The $\beth_{urban}$ for the scenario shown in Figure 6 is about 6°~10° satisfying the sparse area condition.

We can see from the bottom panel of Figure 6, the positioning result of SLAM can well track the ground truth at the beginning of the test. However, due to the accumulated error over time, the SLAM-based trajectory drifts away from the ground truth. The detailed positioning error during the experiment is shown in Figure 7. The top panel shows the positioning error in three different directions. The lateral direction is normal to the driving direction of the vehicle with the longitudinal direction parallel with the driving direction. The bottom panel indicates the reliability estimation. The reliability shown in the bottom panel is calculated based on $C_p\sqrt{U_{ij}}$ presented in Section 2. We can see that the 3D positioning error ($\epsilon_{SLAM}$) almost increases over time with the final positioning error reaching almost 10 meters. The estimated reliability can track the 3D positioning error at the very beginning of the test. The estimated reliability tends to fluctuate between 7 meters over time in this experimental scenario.

Table 2 shows the mean error and standard deviation of positioning error in three separate directions. 3.44 meters of mean error in the lateral direction is obtained with a standard deviation of 1.88 meters. The mean positioning error in the longitudinal direction is 3.19 meters. Moreover, the mean positioning error in altitude direction is just 3.05 meters with a standard deviation of 1.02 meters. The mean of 2D (sum of lateral and longitudinal direction) positioning error is 6.64 meters which are slightly smaller than the 3D positioning error (9.69 meters). As the positioning error accumulates over time using standalone NDT-based graph SLAM, we propose to use the gradient of accumulated error to evaluate the performance of SLAM. The 2D gradient indicates the rate of change of 2D mean positioning error and is obtained with total 2D mean error divided by total epochs. Similarly, the 3D gradient indicates the rate of change of 3D positioning error. The 2D gradient is 0.017, which means that the accumulated error of SLAM increased by meters per second. The 3D positioning error gradient is 0.024 meters per second.

Regarding to the reliability estimation result (blue dots in the bottom panel of Figure 7), we can find that the $C_p\sqrt{U_{ij}}$ overestimates the 3D positioning error in a majority of the time. Moreover, the estimated reliability fluctuates dramatically during the test (from 0 to 15 meters).

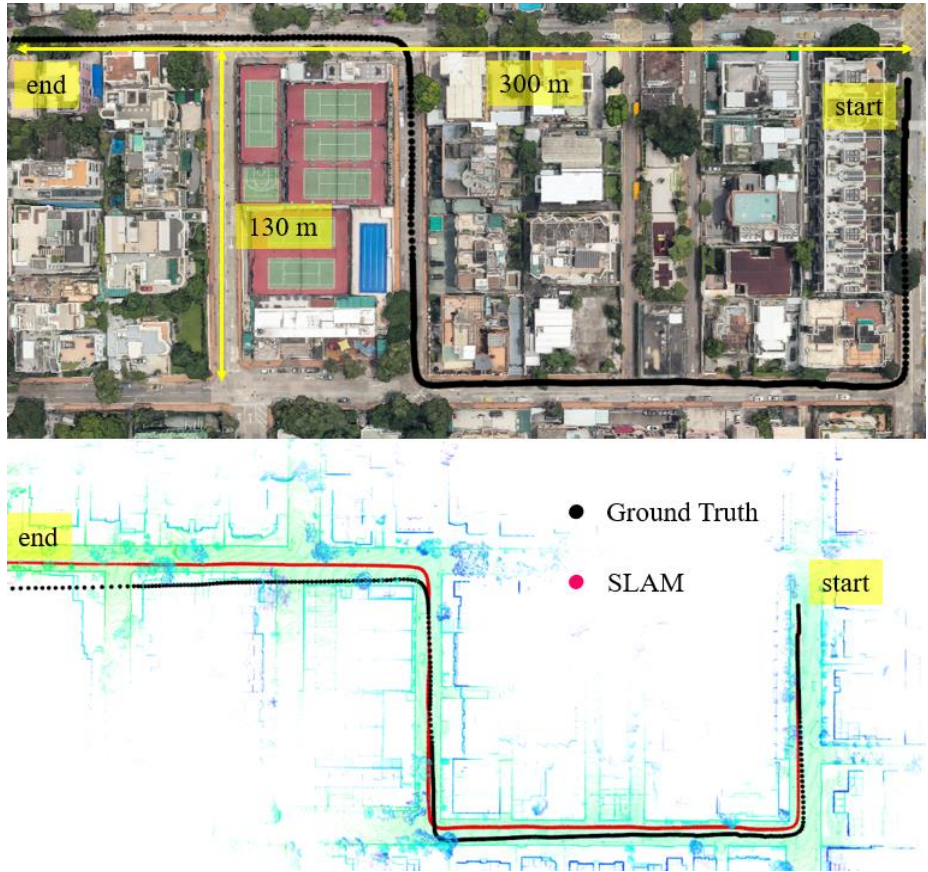

**Figure 6.** Experiment 1: the trajectory of NDT-based graph SLAM in a sparse area with normal traffic condition. Top panel represents the snapshot in Google Maps. The black curve indicates the ground truth of the vehicle's trajectory. The bottom panel indicates the generated points map and trajectory from SLAM and ground truth.

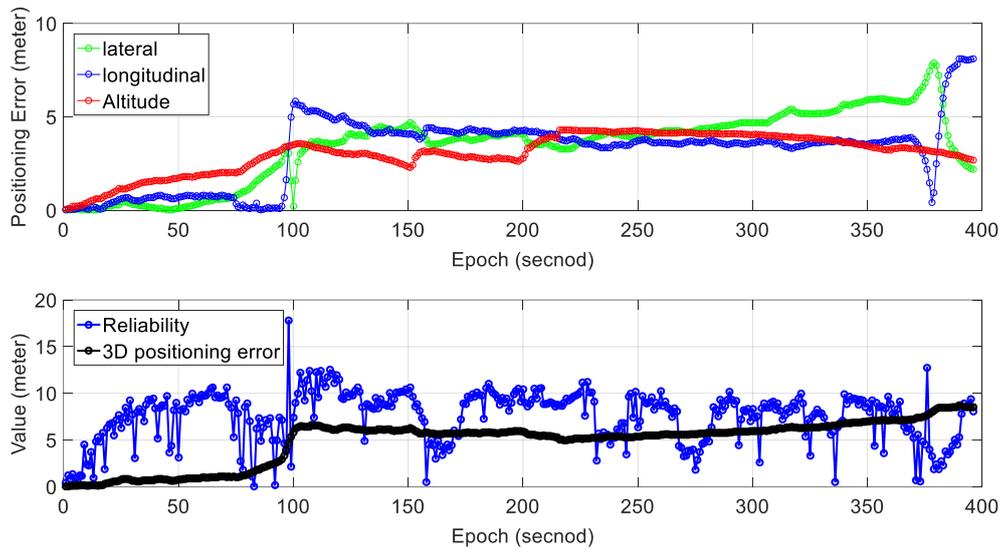

**Figure 7.** Experiment 1: positioning error and reliability estimation result. The top panel represents the positioning error in lateral, longitudinal and altitude directions, respectively. The bottom panel represents the estimated reliability and 3D positioning error of SLAM.

**Table 2.** Experiment 1: Performance of NDT-based graph SLAM in a sparse area with dense traffic condition.

| Error | Lateral (m) | Longitudinal (m) | Altitude (m) | Reliability (m) | 2D (m) | 2D Gradient (m/s) | 3D (m) | 3D Gradient(m/s) |
|---|---|---|---|---|---|---|---|---|
| **Mean** | 3.44 | 3.19 | 3.05 | 7.52 | 6.64 | 0.017 | 9.69 | 0.024 |
| **Std** | 1.88 | 1.79 | 1.02 | 2.75 | 3.29 | 0.008 | 4.12 | 0.010 |

4.2.2 Experiment 2: Performance Evaluation of NDT-based Graph SLAM in Sparse Area with Dense Traffic

In this experiment, the scenario is shown in the top panel of Figure 8. The overall drive of vehicle lasts about 400 seconds in a sparse area with dense traffic. The height of the surrounding buildings is about 5~10 meters high and the width of the streets is approximately 16 meters which are same to experiment 1.

We can see from the bottom panel of Figure 8, the positioning result of SLAM can well track the ground truth at the beginning of the test. However, due to the accumulated error over time, the SLAM-based trajectory is drifting away from the ground truth. The positioning error during the experiment is shown in Figure 9. We can see that the 3D positioning error increases over time with the final positioning error reaching about 19 meters. The estimated reliability can track the 3D positioning error at the very beginning of the test. However, the difference between the estimated reliability of SLAM and the 3D positioning error increases over time. The estimated reliability tends to fluctuate in the vicinity of 5 meters over time during the experiment.

Table 3 shows the mean and standard deviation of positioning error in three separate directions. 6.31 meters of mean error in the lateral direction is obtained with a standard deviation of 5.26 meters. The mean positioning error in the longitudinal direction is 4.91 meters. Interestingly, the mean positioning error in altitude direction is just 0.77 meters with a standard deviation of 0.84 meters. The mean of 2D (sum of lateral and longitudinal directions) positioning error is 11.21 meters which are slightly smaller than the 3D positioning error (11.99 meters). The 2D gradient is 0.028, which means that the accumulated error of SLAM increased by meters per second. The 3D positioning error gradient is 0.03 meters per second.

Regarding to the reliability estimation result (blue dots in the bottom panel of Figure 9), we can find that the estimated reliability is even worse than that in experiment 1. From epoch 100 to 400, the actual 3D positioning error is larger than the estimated (reliability). The estimated mean reliability is 5.93 meters which are smaller than its ground truth (11.99 meters).

The main reason for this is due to the increased traffic density, comparing with experiment 1. According to [40], the dynamic objects from traffic can cause increased uncertainty in LiDAR-based positioning. The evaluated reliability estimation method [34] cannot model the uncertainty caused by dynamic objects. This experiment results show that the traffic has a negative effect on the performance of NDT-based graph SLAM.

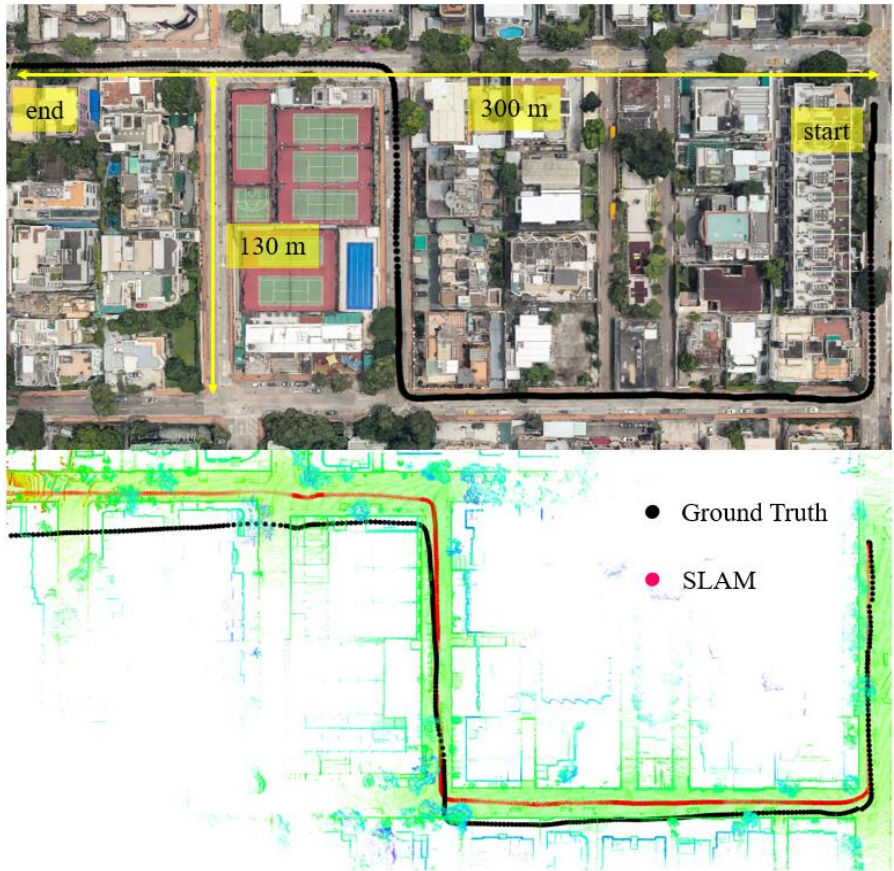

**Figure 8.** Experiment 2: the trajectory of NDT-based graph SLAM in a sparse area with dense traffic condition. Top panel represents the snapshot in Google Maps. The black curve indicates the ground truth of the vehicle's trajectory. The bottom panel indicates the generated points map and trajectory from SLAM.

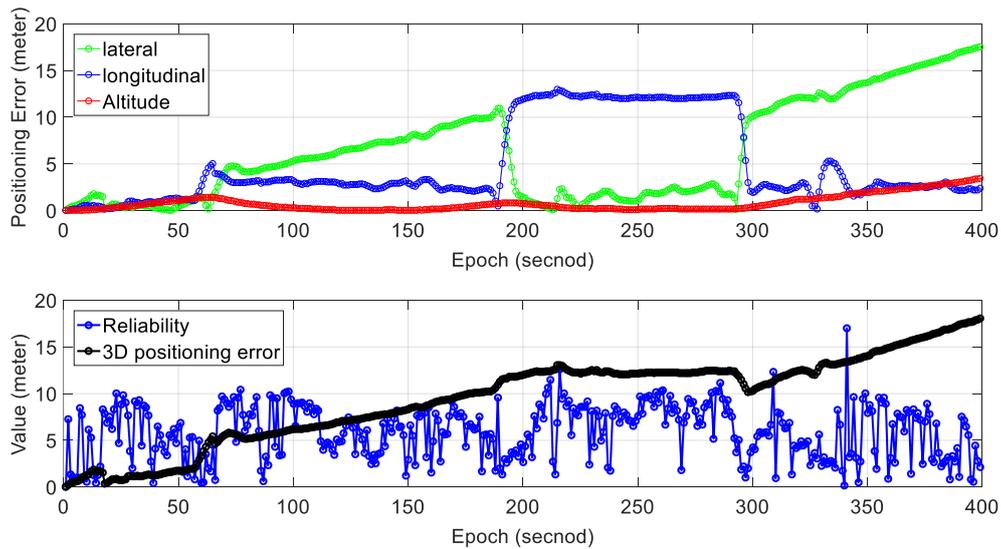

**Figure 9.** Experiment 2: positioning error and reliability estimation result. The top panel represents the positioning error in lateral, longitudinal and altitude directions, respectively. The bottom panel represents the estimated reliability and 3D positioning error of SLAM.

**Table 3.** Experiment 2: Performance of NDT-based graph SLAM in a sparse area with dense traffic condition.

| Error | Lateral (m) | Longitudinal (m) | Altitude (m) | Reliability (m) | 2D (m) | 2D Gradient (m/s) | 3D (m) | 3D Gradient(m/s) |
|---|---|---|---|---|---|---|---|---|
| **Mean** | 6.31 | 4.91 | 0.77 | 5.93 | 11.21 | 0.028 | 11.99 | 0.03 |
| **Std** | 5.26 | 4.36 | 0.84 | 2.87 | 5.18 | 0.013 | 5.60 | 0.014 |

*4.3 Experiment in Dense Urban Area*

4.3.1 Experiment 3: Performance Evaluation of NDT-based Graph SLAM in Dense Urban Area with Normal Traffic

In this experiment, the scenario is shown in the top panel of Figure 10. The overall drive of vehicle lasts about 124 seconds in dense urban with normal traffic. The height of the surrounding buildings is about 59~105 meters high and the width of the streets is approximately 16~20 meters. The $\beth_{urban}$ for the scenario shown in Figure 10 is about 50°~63° satisfying the dense urban area condition.

We can see from the bottom panel of Figure 11, the positioning result of SLAM can well track the ground truth at the beginning of the test. However, due to the accumulated error over time, the SLAM-based trajectory is drifting away from the ground truth. The positioning error during the experiment is shown in Figure11. We can find that the main trend in positioning error is that the positioning error $\epsilon_{SLAM}$ increased over time. The 3D positioning error increases over time with the final positioning error reaching about 28 meters. The estimated reliability can track the 3D positioning error at the very beginning of the test. However, the difference between the estimated reliability of SLAM and the 3D positioning error increases over time. The estimated reliability tends to fluctuate between 10 and 60 meters over time during the experiment.

Table 4 shows the mean and standard deviation of positioning error in three separate directions. 6.73 meters of mean error in the lateral direction is obtained with a standard deviation of 4.39 meters. The mean positioning error in the longitudinal direction is 4.81 meters. Interestingly, the mean positioning error in altitude direction is just 11.90 meters which are significantly larger than the other two directions. The mean of 2D (sum of lateral and longitudinal directions) positioning error is 11.54 meters with the 3D positioning error reaching 14.85 meters. The 2D gradient is 0.094 and the value for the 3D gradient is 0.121.

Regarding the reliability estimation result (blue dots in the bottom panel of Figure 11), from epoch 0 to 60, the actual 3D positioning error is smaller than the estimated (reliability). The estimated mean reliability is 18.90 meters which are significantly larger than its ground truth (14.85 meters).

By comparing with the experiments conducted in the sparse area, the mean 3D gradient of 3D positioning error in dense urban (0.121 m/s) is significantly larger than that in the sparse area (0.024 m/s in normal traffic and 0.03 in normal traffic). These results show that the degree of urbanization also has an impact on the performance of NDT-based graph SLAM.

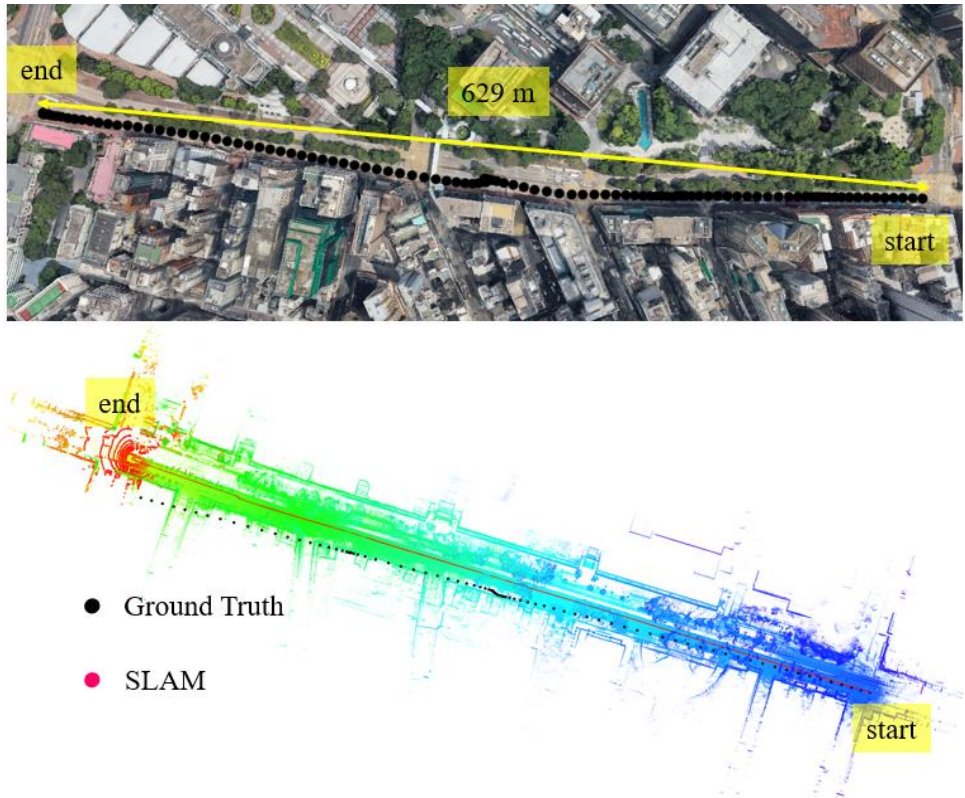

**Figure 10.** Experiment 3: the trajectory of NDT-based graph SLAM in a dense urban area with normal traffic condition. Top panel represents the snapshot in Google Maps. The black curve indicates the ground truth of the vehicle's trajectory. The bottom panel indicates the generated points map and trajectory from SLAM.

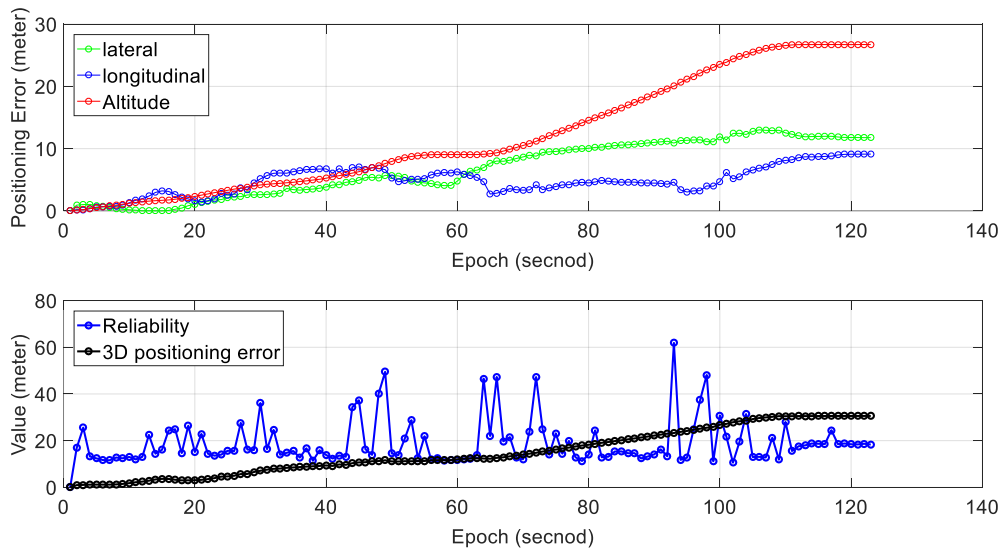

**Figure 11.** Experiment 3: positioning error and reliability estimation result. The top panel represents the positioning error in lateral, longitudinal and altitude directions, respectively. The bottom panel represents the estimated reliability and 3D positioning error of SLAM.

**Table 4.** Experiment 3: Performance of NDT-based graph SLAM in a dense urban area with normal traffic condition.

| Error | Lateral (m) | Longitudinal (m) | Altitude (m) | Reliability (m) | 2D (m) | 2D Gradient (m/s) | 3D (m) | 3D Gradient(m/s) |
|---|---|---|---|---|---|---|---|---|
| **Mean** | 6.73 | 4.81 | 11.9 | 18.9 | 11.54 | 0.094 | 14.85 | 0.121 |
| **Std** | 4.39 | 2.31 | 9.01 | 9.61 | 6.01 | 0.049 | 9.75 | 0.079 |

4.3.2 Experiment 4: Performance Evaluation of NDT-based Graph SLAM in Dense Urban Area with Dense Traffic

In this experiment, the scenario is shown in the top panel of Figure 12. The overall drive of vehicle lasts about 124 seconds in dense urban with dense traffic. The height of the surrounding buildings is about 50~175 meters high and the width of the streets is approximately 16~20 meters. The $ℶ_{urban}$ for the scenario shown in Figure 10 is about 50°~67° satisfying the dense urban area condition.

We can see from the bottom panel of Figure 13, the positioning result of SLAM can well track the ground truth at the beginning of the test. However, due to the accumulated error over time, the SLAM-based trajectory is drifting away from the ground truth. The positioning error ($\epsilon_{SLAM}$) during the experiment is shown in Figure13. The 3D positioning error increases over time with the final positioning error reaching about 42 meters. The estimated reliability can track the 3D positioning error at the very beginning of the test. However, the difference between the estimated reliability of SLAM and the 3D positioning error increases over time. Interestingly, we can find that altitude positioning error in both experiment 3 and experiment 4 dominant the $\epsilon_{SLAM}$.

Table 5 shows the mean and standard deviation of positioning error in three separate directions. 11.25 meters of mean error in the lateral direction is obtained with a standard deviation of 5.09 meters. The mean positioning error in the longitudinal direction is 2.77 meters. Interestingly, the mean positioning error in altitude direction reaches 19.07 meters which are significantly larger than the other two directions. The mean of 2D (sum of lateral and longitudinal directions) positioning error is 14.02 meters with the 3D positioning error reaching 23.22 meters. The 2D gradient is 0.114 and the value for the 3D gradient is 0.189.

Regarding the reliability estimation result (blue dots in the bottom panel of Figure 13), from epoch 20 to 120, the actual 3D positioning error is larger than the estimated (reliability). The estimated mean reliability is 14.38 meters which are significantly smaller than its ground truth (23.22 meters). Interestingly, we can find that it tends to overestimate the uncertainty of SLAM in normal traffic scenario and underestimate that in dense traffic scene. The main reason for this is that the used reliability estimation method cannot model the effects of traffic (dynamic objects). In other words, the dense traffic scenes introduce larger uncertainty. This result again shows that the traffic has a bad impact on the performance of NDT-based graph SLAM.

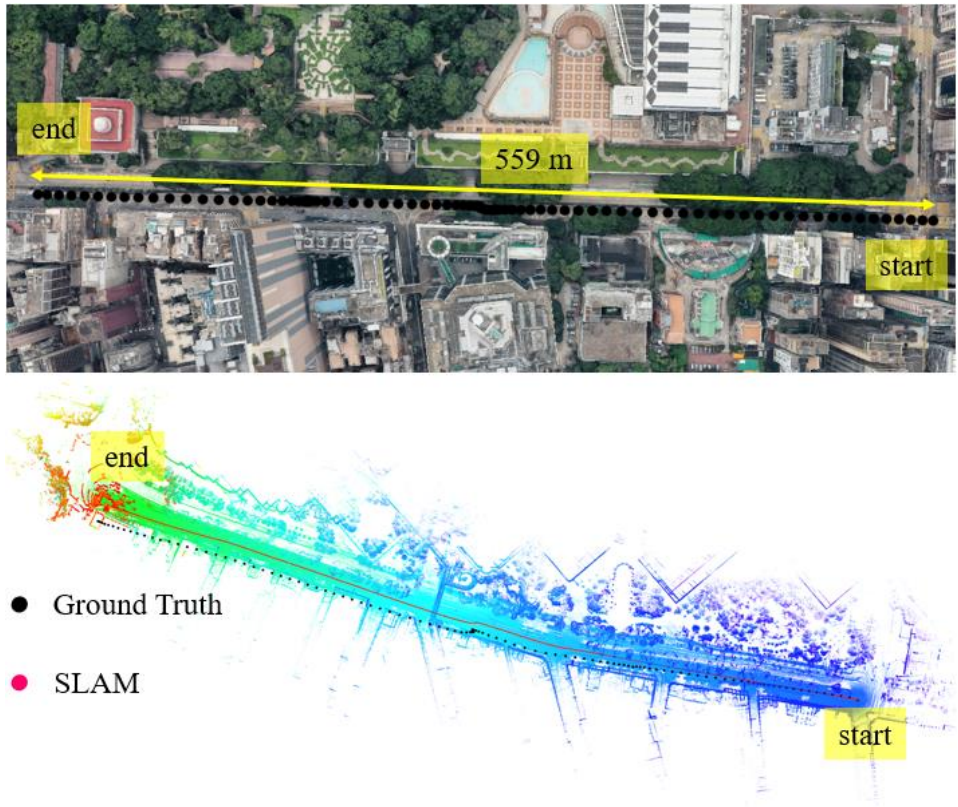

**Figure 12.** Experiment 4: the trajectory of NDT-based graph SLAM in a dense urban area with dense traffic condition. Top panel represents the snapshot in Google Maps. The black curve indicates the ground truth of the vehicle's trajectory. The bottom panel indicates the generated points map and trajectory from SLAM.

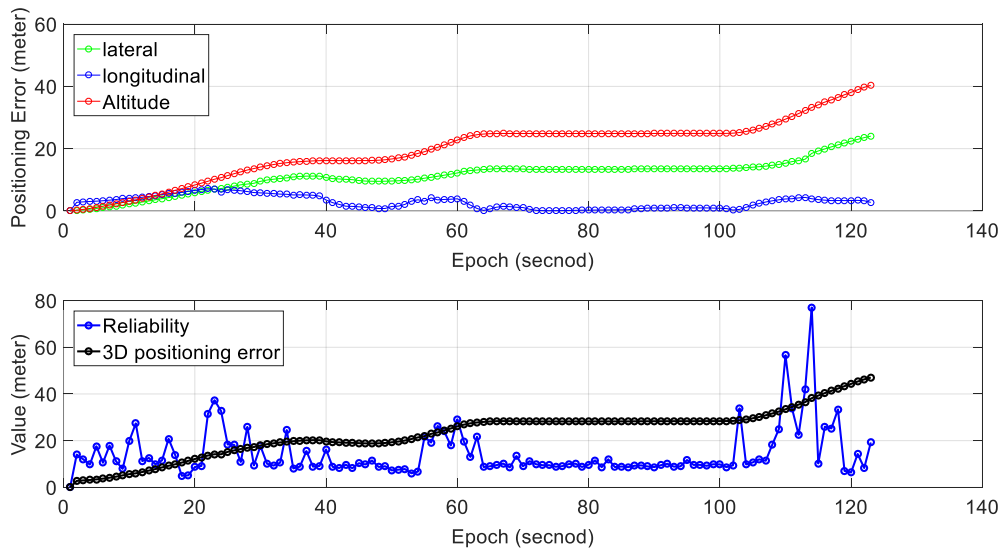

**Figure 13.** Experiment 4: positioning error and reliability estimation result. The top panel represents the positioning error in lateral, longitudinal and altitude directions, respectively. The bottom panel represents the estimated reliability and 3D positioning error of SLAM.

**Table 5.** Experiment 4: Performance of NDT-based graph SLAM in dense urban area with dense traffic condition.

| Error | Lateral (m) | Longitudinal (m) | Altitude (m) | Reliability (m) | 2D (m) | 2D Gradient (m/s) | 3D (m) | 3D Gradient(m/s) |
|---|---|---|---|---|---|---|---|---|
| Mean | 11.25 | 2.77 | 19.07 | 14.38 | 14.02 | 0.114 | 23.22 | 0.189 |
| Std | 5.09 | 2.06 | 9.60 | 10.25 | 4.80 | 0.039 | 10.25 | 0.083 |

## 5. Discussion, Conclusion and Future Work

(1) The relationship between the traffic conditions and the performance of NDT-based graph SLAM positioning:

**Traffic condition and accuracy of NDT-based graph SLAM**: The detailed analysis of the relationship between the traffic conditions and the accuracy of LiDAR-based positioning is shown in Figure 14 which shows the results in two different degrees of traffic conditions. According to the presented 6 experiments (including 2 experiments presented in the Appendix); the accuracy of the SLAM is degraded with increased traffic density. For example, the mean 3D positioning error increased from 1.58 meters (experiment 5 with normal traffic) to 1.91 meters (experiment 6 with dense traffic). This phenomenon is also the same in the sparse area and dense urban areas. The main reason causing this degradation in SLAM performance is the moving objects in traffic, such as the double-decker bus, cars, and trucks. Our previous research [20] shows that the height of the double-decker bus can go up to 4.5 meters in Hong Kong and thus take up the majority of the field of view (FOV) of 3D LiDAR. The double-decker bus is a moving object on the roads. In this case, the majority of the 3D point clouds are scanned from the moving objects. The points from moving objects can distort the mapping between two consecutive frames of point clouds, thus impairing the accuracy of SLAM. We can also see that the positioning error gradient increases with enhanced traffic density. In overall, the traffic condition has bad effects on the accuracy of NDT-based SLAM. In other words, the more dynamic environments with more moving objects introduce more degradation in the positioning accuracy of NDT-based graph SLAM. The evaluated results related to the traffic and accuracy of NDT-based SLAM can be a good benchmark for further mitigating the effects of traffic to improve the accuracy of LiDAR-based positioning.

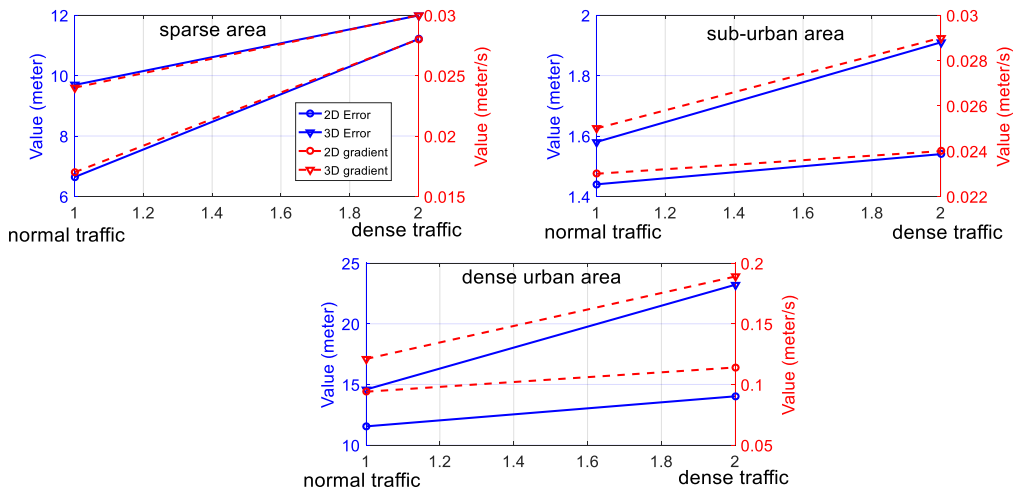

**Figure 14.** The relationship between the traffic conditions and the performance of NDT-based graph SLAM. The x-axis indicates the traffic condition (including normal traffic and dense traffic). The y-axis on the left side

represents the value of positioning error and the y-axis on the right side indicates the positioning error gradient.

**Traffic condition and reliability estimation**: As presented in Experiment 4, it tends to overestimate the uncertainty of SLAM in normal traffic scenario and underestimate that in dense traffic scene. In other words, the dense traffic scenes introduce larger uncertainty. This result again shows that the traffic has a bad impact on the performance of NDT-based graph SLAM. Proper methods to cope with the dynamic objects are needed to effectively estimate the uncertainty caused by dense traffic.

(2) The relationship between the degree of urbanization and the performance of LiDAR-based positioning:

**The degree of urbanization and accuracy of NDT-based graph SLAM**: The detailed analysis of the relationship between the degree of urbanization and the accuracy of NDT-based graph SLAM is shown in Figure 15. According to the 6 experiments, three levels of areas classified based on the degree of urbanization are presented. We can see from Figure 15, the 3D gradient in sub-urban is similar to that in the sparse area. However, the 3D gradient in dense urban is significantly larger than that in both sparse areas and sub-urban. The main reason for this result is the environment features availability. In the sub-urban and sparse areas experiments, the main features are buildings, moving objects, and some trees, which means that abundant features are available. In the dense urban area, the main features are tall buildings and moving objects, which means less feature availability. Moreover, we can find that the 3D positioning error in altitude direction increases dramatically with an increased degree of urbanization which can be seen by comparing with Experiment 1 and 3. In total, the increased density of urbanization can degrade the accuracy of SLAM-based positioning, especially in the altitude direction. To effectively model the uncertainty of LiDAR-based positioning, the surrounding environment features are needed to be considered, for example, the degree of urbanization. The 3D building model generated in Figure 2 is a potential resource which contains the models of buildings to improve the accuracy of NDT-based graph SLAM. Inspired by this, we are going to employ the 3D building model to facilitate the effects estimation of urbanization on the performance of LiDAR-based positioning.

**The degree of urbanization and reliability estimation of NDT-based graph SLAM**: as discussed in experiment 4, the reliability estimation of NDT-based SLAM is highly related to the traffic condition. However, there are no obvious relations between the reliability estimation and the degree of urbanization according to the presented experiments. In total, the sub-urban area has the smallest mean uncertainty and dense urban possesses the largest estimated uncertainty.

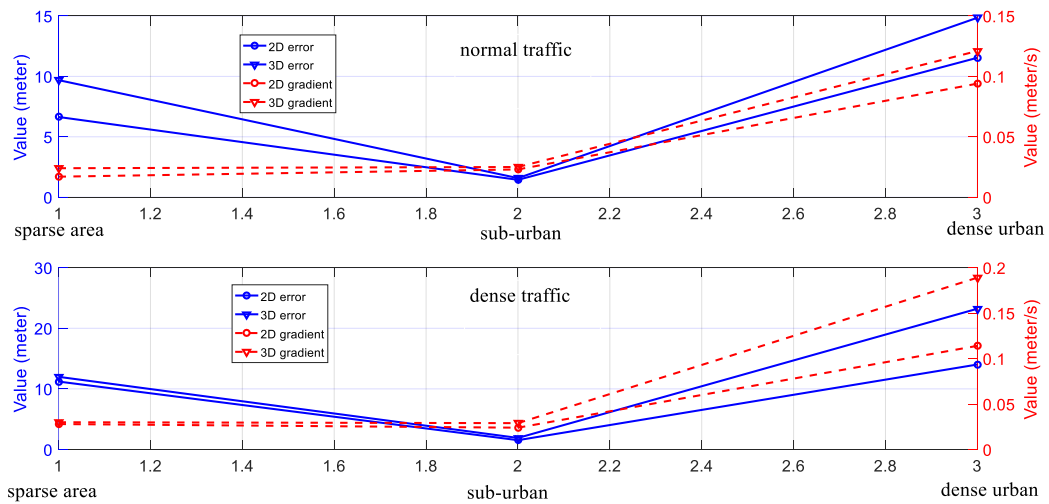

**Figure 15.** The relationship between the degree of urbanization and the performance of LiDAR-based positioning. The x-axis indicates the degree of urbanization. The y-axis on the left side represents the value of positioning error and the y-axis on the right side indicates the positioning error gradient.

Insufficient positioning accuracy and robustness is one of the main technic problems that prevent the arrival of autonomous vehicles in the super-urbanized areas, such as Hong Kong, Tokyo, and New York. According to our previous localization experiments conducted in Hong Kong using multi-sensor fusion (GNSS/IMU/LiDAR/HD Map) method [34] using Kalman filter, we find that the method (GNSS/IMU/LiDAR/HD Map) can obtain satisfactory performance in the sparse area. However, the solution usually fails in a deep urban area with dense traffic. Therefore, we propose to conduct experiments in diverse urban scenarios to find out the effects of traffic and degree of urbanization on LiDAR-based positioning. This is also the main contribution of this paper. Coping with the effects from both the traffic and urbanization on the LiDAR-based positioning is significant for the GNSS/INS/LiDAR/HD Map-based localization solution for an autonomous vehicle. We believe that this paper can be a useful basic work to improving the LiDAR-based positioning in dense urban scenarios.

(3) **Future work:** the moving objects detection and 3D building models will be employed to improve the performance of NDT-based graph SLAM. Moreover, the uncertainty estimation of LiDAR-based SLAM will be conducted by considering both the traffic conditions and 3D building models.

**Acknowledgments:** The authors acknowledge the support of Hong Kong PolyU startup fund on the project 1-ZVKZ, "Navigation for Autonomous Driving Vehicle using Sensor Integration".

**Author Contributions:** Conceptualization, Li-ta Hsu; Investigation, Weisong WEN; Methodology, Guohao Zhang.

**Conflicts of Interest:** The authors declare no conflict of interest.

**Appendix**

*Experiment in Sub-urban Area*

Experiment 5: Performance Evaluation of NDT-based Graph SLAM in Sub-urban Area with Normal Traffic

In this experiment, the scenario is shown in the right panel of Figure 16. The overall drive of vehicle lasts about 64 seconds in sub-urban with dense traffic. The height of the surrounding buildings is about 10~25 meters high and the width of the streets is approximately 16~20 meters. The $\beth_{urban}$ for the scenario shown in Figure 16 is about 20°~41° satisfying the dense urban area condition.

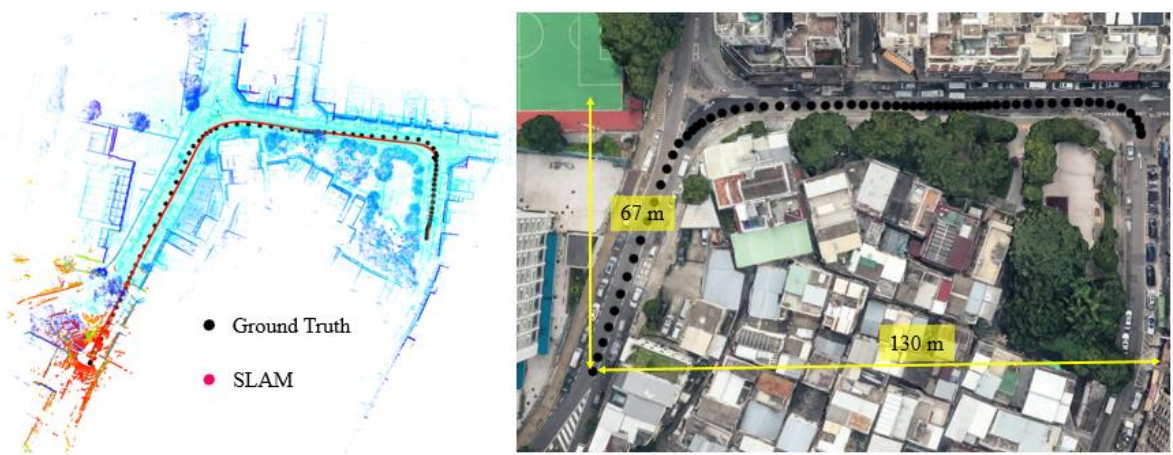

**Figure 16.** Experiment 5: the trajectory of NDT-based graph SLAM in a sub-urban area with normal traffic condition. The left panel indicates the generated points map and trajectory from SLAM. Right panel represents the snapshot in Google Maps. The black curve indicates the ground truth of the vehicle's trajectory.

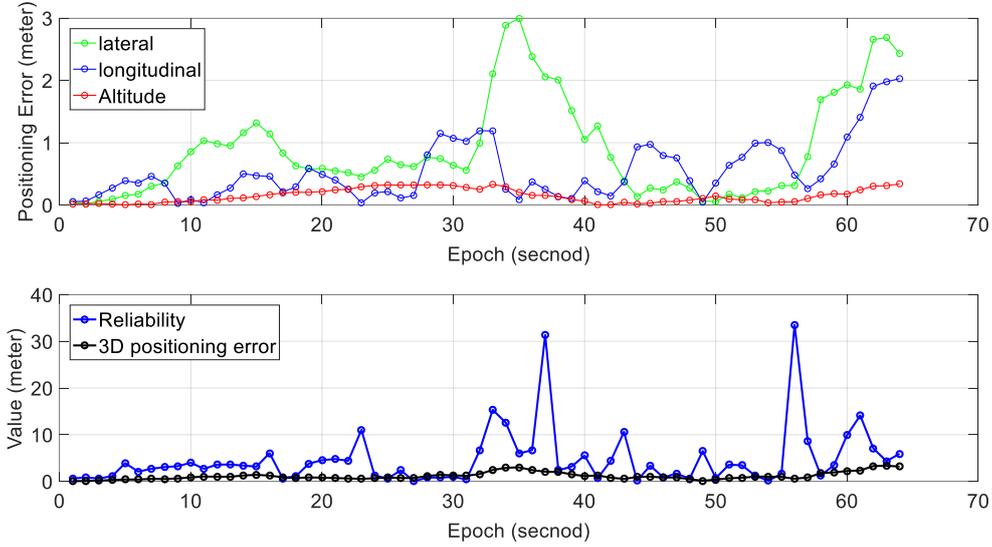

**Figure 17.** Experiment 5: positioning error and reliability estimation result. The top panel represents the positioning error in lateral, longitudinal and altitude directions, respectively. The bottom panel represents the estimated reliability and 3D positioning error of SLAM.

**Table 6.** Experiment 5: Performance of NDT-based graph SLAM in a sub-urban area with normal traffic condition.

| Error | Lateral (m) | Longitudinal (m) | Altitude (m) | Reliability (m) | 2D (m) | 2D Gradient (m/s) | 3D (m) | 3D Gradient(m/s) |
|---|---|---|---|---|---|---|---|---|
| **Mean** | 0.91 | 0.54 | 0.15 | 4.69 | 1.44 | 0.023 | 1.58 | 0.025 |
| **Std** | 0.79 | 0.48 | 0.11 | 6.09 | 1.05 | 0.016 | 1.11 | 0.016 |

Experiment 6: Performance Evaluation of NDT-based Graph SLAM in Sub-urban Area with Dense Traffic

In this experiment, the scenario is shown in the right panel of Figure 18. The overall drive of vehicle lasts about 64 seconds in sub-urban with dense traffic. The height of the surrounding buildings is about 10~25 meters high and the width of the streets is approximately 16~20 meters. The ℶ$_{urban}$ for the scenario shown in Figure 18 is about 20°~41° satisfying the dense urban area condition. The only difference between experiment 5 and experiment 6 is the density of traffic.

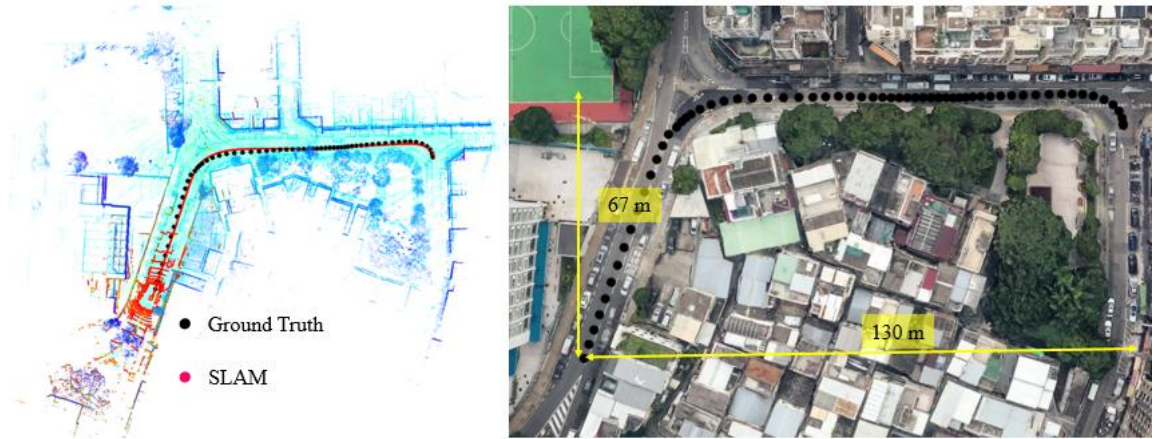

**Figure 18.** Experiment 6: the trajectory of NDT-based graph SLAM in a sub-urban area with dense traffic condition. The left panel indicates the generated points map and trajectory from SLAM. Top panel represents the snapshot in Google Maps. The black curve indicates the ground truth of the vehicle's trajectory.

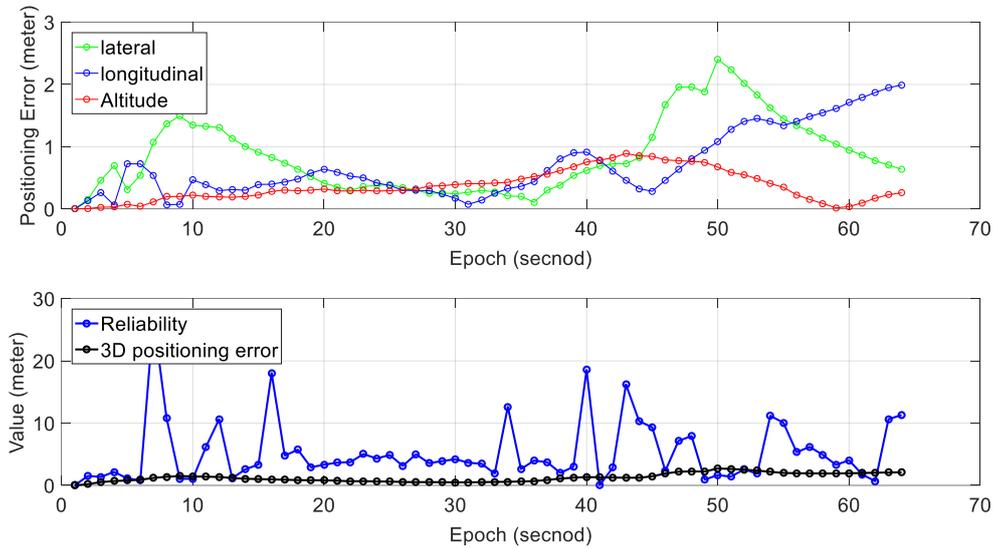

**Figure 19.** Experiment 6: positioning error and reliability estimation result. The top panel represents the positioning error in lateral, longitudinal and altitude directions separately. The bottom panel represents the estimated reliability and 3D positioning error of SLAM.

**Table 7.** Experiment 4: Performance of NDT-based graph SLAM in a sub-urban area with dense traffic condition.

| Error | Lateral (m) | Longitudinal (m) | Altitude (m) | Reliability (m) | 2D (m) | 2D Gradient (m/s) | 3D (m) | 3D Gradient(m/s) |
|---|---|---|---|---|---|---|---|---|
| **Mean** | 0.85 | 0.694 | 0.367 | 5.26 | 1.54 | 0.024 | 1.91 | 0.029 |
| **Std** | 0.59 | 0.537 | 0.25 | 5.05 | 0.94 | 0.015 | 1.01 | 0.016 |